\documentclass[lettersize,journal]{IEEEtran}
\usepackage{amsmath,amsfonts}
\usepackage{algorithmic}
\usepackage{algorithm}
\usepackage{array}
\usepackage[caption=false,font=normalsize,labelfont=sf,textfont=sf]{subfig}
\usepackage{textcomp}
\usepackage{stfloats}
\usepackage{url}
\usepackage{verbatim}
\usepackage{graphicx}
\usepackage{cite}
\usepackage{makecell}
\usepackage{multirow}
\usepackage{url}
\usepackage[backref]{hyperref}
\usepackage{xcolor}

\begin{document}

\title{Diffusion-based Graph Generative Methods}

\author{Hongyang Chen,~\IEEEmembership{Senior member,~IEEE,} Can Xu, Lingyu Zhen, Qiang Zhang, Xuemin Lin,~\IEEEmembership{Fellow,~IEEE,}
\IEEEcompsocitemizethanks{
    \IEEEcompsocthanksitem Correspond to Hongyang Chen.
    \IEEEcompsocthanksitem Hongyang Chen is with the Research Center for Graph Computing, Hangzhou, China. Email: hongyang@zhejianglab.com.
    \IEEEcompsocthanksitem Can Xu is with School of Data Science and Engineering, East China Normal University, Shanghai, China, and also with the Research Center for Graph Computing, Hangzhou, China. Email: leoxc1571@163.com.
    \IEEEcompsocthanksitem Lingyu Zheng is with Hangzhou Institue for Advanced Study, UCAS, Hangzhou, China. Email: zhenglingyu22@mails.ucas.ac.cn.
    \IEEEcompsocthanksitem Qiang Zhang is with the College of Computer Science and Technology, Zhejiang University, Hangzhou, China, and also with ZJU-Hangzhou Global Scientific and Technological Innovation Center, Hangzhou, China. Email: qiang.zhang.cs@zju.edu.cn.
    \IEEEcompsocthanksitem Xuemin Lin is with Antai College of Economics and Management, Shanghai Jiao Tong University, Shanghai, China. Email: xuemin.lin@sjtu.edu.cn.
    \IEEEcompsocthanksitem Hongyang Chen and Can Xu contributed equally to this work. \protect\\
}

\thanks{}}

\markboth{Journal of \LaTeX\ Class Files,~Vol.~14, No.~8, August~2021}%
{Shell \MakeLowercase{\textit{et al.}}: A Sample Article Using IEEEtran.cls for IEEE Journals}


\maketitle

\begin{abstract}
    Being the most cutting-edge generative methods, diffusion methods have shown great advances in wide generation tasks. Among them, graph generation attracts significant research attention for its broad application in real life. In our survey, we systematically and comprehensively review on diffusion-based graph generative methods. We first make a review on three mainstream paradigms of diffusion methods, which are denoising diffusion probabilistic models, score-based genrative models, and stochastic differential equations. Then we further categorize and introduce the latest applications of diffusion models on graphs. In the end, we point out some limitations of current studies and future directions of future explorations. The summary of existing methods metioned in this survey is in our Github: \href{https://github.com/zhejiangzhuque/Diffusion-based-Graph-Generative-Methods}{https://github.com/zhejiangzhuque/Diffusion-based-Graph-Generative-Methods}.
\end{abstract}

\begin{IEEEkeywords}
    Generative methods, diffusion models, graph neural networks, molecule generation, motion generation.
\end{IEEEkeywords}

\section{Introduction}
\IEEEPARstart{D}{eep} generative methods, especially diffusion-based methods, have shown promising potentials in a variety of domains. Notably, generative methods such as generative adversarial networks (GANs) \cite{gan_14_goodfellow,conditionalgan_14_mirza,wgan_17_arjovsky}, variational autoencoders (VAEs) \cite{vae_13_kingma,vae_14_rezende,tutorial_16_doersch,introvae_19_kingma}, auto-regressive models (ARs) \cite{ar_16_oord}, normalizing flows (NFs) \cite{nice_15_dinh,density_17_dinh}, and diffusion-based models \cite{dpm_15_jascha,sgm_19_song,ddpm_20_ho,scoresde_21_song}, have shown to be capable of creating novel examples that are challenging to distinguish for humans. Recent advances in diffusion-based methods show promising performance in generating various kinds of samples compares to previous generative methods. Among them, graph generation has wide applications in multiple fields, such as computational chemistry \cite{moflow_20_zang,graphaf_20_shi,genchemical_23_dylan} and character animation \cite{motiongraph_12_min}. This survey aims to provide a comprehensive overview of the cutting-edge applications and developments of diffusion-based graph generative methods. Compares to previous survey \cite{graphdiffsurvey_23_zhang,gendiffgraph_23_liu}, this work provides a more structural overview on diffusion-based generative methods on all kinds of graph data. To the best of our knowledge, this work is the most exhaustive survey on diffusion-based graph generative methods so far. 

Graphs are natural medium for modeling real-life macro and micro objects. By mapping real-life objects to vertices and edges, graphs are capable to represent unstructured data. Applications of them range from social networks, recommender systems \cite{dscf_19_fan,ngcf_10_wang}, to molecular learning, \cite{cgcnn_18_xie,attentivefp_20_xiong,gnnmol_23_guo} and et al. Previous methods have achieved great success in graph generation tasks. Auto-regressive methods \cite{gspherenet_22_luo,3dgensbdd_21_luo} are able to generate nodes and edges one by one based on given information and former generated ones, which may lead to early stopping and ignorance of global interactions. Some one-shot generation methods, such as VAEs and GANs \cite{dmcg_22_zhu,ligan_20_masuda,graphdg_20_simm}, significantly relieve problems of sequential generation methods. However, the end-to-end structure of these methods are hard to train compare to diffusion methods, which are capable to minimize differences between predictions and inputs at each time step.

With the advent of diffusion models \cite{ddpm_20_ho,dpm_15_jascha,sgm_19_song,scoresde_21_song}, their applications on graphs arouse significant scientific interests. While majority of diffusion-based models operate on continuous data, molecule design combined with 3D structure \cite{edm_22_hoogeboom,diffbp_22_lin} and 3D motion synthesis \cite{motiondiffuse_22_zhang,modiff_23_zhao} rapidly emerged. However, due to the intrinsic discrete nature of graphs, there are a number of significant challenges when implementing diffusion models to graph generation. Since graphs widely exist in heterogeneous forms in practical applications, discretization of features as well as following certain rules and dependencies of graph structures remain major challenges. In this paper, we primarily concentrate on introducing the most advanced research in generative diffusion methods on graph data. Among various graph generation tasks, diffusion-based methods are widely adopted in molecule generation and motion generation. To tackle the challenges posed by the discrete nature of graphs, many research efforts are devoted to adapting diffusion models to discrete data generation \cite{digress_22_vignac,graphgdp_22_huang,nvdiff_22_chen}. 

In terms of the outline of this survey, we first elaborate on mainstream paradigms of diffusion methods and their applications on graphs. Furthermore, we make a systematic and structured review on applications of generative diffusion methods in molecule generation and motion generation. Over recent years, the incorporation of 3D structure have brought a renaissance in molecular learning research. In the field of 3D molecule design, diffusion-based methods are capable of generating elements like molecules, conformations, ligands, linkers, proteins, amongst others. In terms of motion synthesis, research shows diffusion models are capable to generate long-sequence motion from conditonal information or sequence. Next, we summarize all the datasets and evaluation metrics employed in mentioned applications. In the end, we make a bief summary on current methods and an outlook on possible future directions.

\section{Problem formulation}
\begin{figure*}[t]
    \centering
    \includegraphics[width=0.7\linewidth]{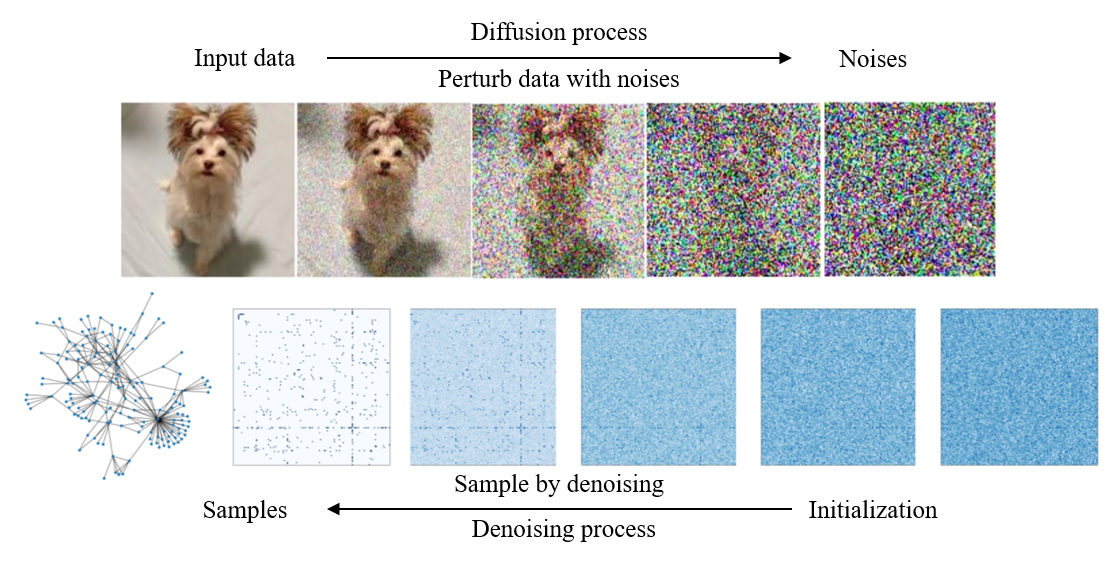}
    \caption{An illustration of diffusion models on images and graphs \cite{scoresde_21_song,gsdm_22_luo}. The right arrow points at the direction of diffusion process, where noises are injected to ground-truth data. The left one indicates the sampling phase where samples are generated.}
    \label{fig:gsdm}
\end{figure*}

In the realm of graph learning, a graph is formally defined as a tuple $\mathcal{G} = (\mathcal{V}, \mathcal{E})$, cosisting of a vertex set $\mathcal{V}$ and an edge set $\mathcal{E}$. During the training period, the probabilistic distirbution of graphs in the diffusing process are learned and are later used in the sampling phases, where new graphs are created iteratively. 

Diffusion-based models inject noises progressively into the original data and subsequently generate samples in the reverse process. Denoising probabilistic models (DDPMs) \cite{ddpm_20_ho,dpm_15_jascha}, score-based generative models (SGMs) \cite{sgm_20_song,sgm_19_song} and stochastic differential equations (Score SDEs) \cite{scoresde_21_song,mlscoresde_21_song} are the three main paradigms of diffusion methods. In Figure~\ref{fig:gsdm}, we present the intuition of diffusion methods for computer vision and graphs. The summary of diffusion-based graph generative methods are also listed in Table~\ref{tab:sum_graphdiff}.

\subsection{Denoising diffusion probabilistic models (DDPMs)}
A denoising diffusion probabilistic model \cite{ddpm_20_ho,dpm_15_jascha} consists of two Markov chains that each serves as the diffusion process and the denoising process. The forward chain iteratively adds noise to the original data and transform it into a prior distribution. During the sampling phase, the reverse chain learns to reverse the diffusion process by gradually denoising the prior distribution into the origin data. 

Given the raw data distribution $x_0 \sim q(x_0)$ and the posterior distribution of noise $q(x_t | x_{t-1})$, the forward chain gradually perturbs $x_0$ into $x_t (t = 1, 2, ..., T)$. Theoretically, the transition distribution $q(x_t | x_{t-1})$ allows customization. In the follow-up discussion, we take the most commonly used Gaussian distribution as an example. With a fixed variance schedule $\beta_t \in (0, 1) (t = 1, 2, ..., T)$, the posterior distribution of noise is
\begin{equation}
    q(x_t | x_{t-1}) = \mathcal{N}(x_t; \sqrt{1-\beta_t} x_{t-1}, \beta_t I).
\end{equation}

According to the Markov property, the joint distribution of $x_1, x_2, ..., x_T$ conditioned on $x_0$ is 
\begin{equation}
    q(x_1, x_2, ..., x_T | x_0) = \prod_{t=1}^T q(x_t | x_{t-1}).
\end{equation}

Let $\alpha_t = 1 - \beta_t$ and $\bar{\alpha}_t = \prod_{s=1}^t \alpha_s$, the posterior distribution of data at any time step $t$ can be calculated $q(x_t | x_0) = \mathcal{N}(x_t; \sqrt{\bar{\alpha}_t} x_0, (1-\bar{\alpha}_t) I)$. Let $\epsilon \sim \mathcal{N}(0, I)$, we have 
\begin{equation}
    x_t = \sqrt{\bar{\alpha}_t} x_0 + \sqrt{1 - \bar{\alpha}_t} \epsilon.
\end{equation}

With adequately large time step, the data will be converted to isotropic Gaussian. which means the distribution of $x_T$ converges to $\mathcal{N}(x_T; 0, I)$.

During the generation phase, a noise vector $x_T$ is drawn from the prior noise distribution $p(x_T)$. Then the denoising process gradually reverses the forward chain with the help of transition kernel $p_{\theta}(x_{t-1}|x_t)$ until reaching the initial time step. The reverse kernel and the distribution of sampling results can be further represented as:
\begin{align}
    p_\theta(x_{t-1} | x_t)  &= \mathcal{N}(x_{t-1}: \mu_\theta(x_t, t), \Sigma_\theta(x_t, t)), &\\
    p_\theta(x_0, x_1, ..., x_T) &:= p(x_T) \prod_{t=1}^T p_\theta(x_{t-1} | x_t), \ p_\theta(x_0) \\
    &= \int p_\theta(x_0, x_1, ..., x_T) dx_1 dx_2 ... dx_T,
\end{align}
where $\mu_\theta(x_t, t)$ and $\Sigma_\theta(x_t, t)$ are learnable deep neural networks and $\theta$ represents model parameters. At the training stage, optimizing parameters enables the reverse chain to mimic the reverted forward chain accurately. Therefore the objective function is formulated as Kullback-Leibler (KL) divergence between the predicting outcomes of deep neural networks and the pre-defined forward chain:
\begin{align}
    &E[KL(q(x_0, x_1, ..., x_T) || p_\theta(x_0, x_1, ..., x_T))] \\
    =& E_q[KL(q(x_T | x_0) || p(x_T)) + \notag \\
    &\sum_{t>1}KL(q(x_{t-1} | x_t, x_0) || p_\theta(x_{t-1} | x_t)) - log p_\theta(x_0 | x_1)] \\
    =& E_q[-log p(x_T) - \sum_{t=1}^T log \frac{p_\theta(x_{t-1} | x_t)}{s(x_t | x_{t-1})}] + const \label{eq:vlb} \\ 
    \geq& E[-log p_\theta(x_0)] + const,
\end{align}
where (\ref{eq:vlb}) is the variational lower bound (VLB) of the log-likelihood of $x_0$, which is a typical target function in DDPMs.

DDPMs have found broad applicability in graph generation tasks, with EDM \cite{edm_22_hoogeboom}, GeoDiff \cite{geodiff_22_xu}, and MotionDiffuse \cite{motiondiffuse_22_zhang} being some of the pioneers in applications of DDPMs. Most research focuses on the innovation of the design of denoising kernel. EDM \cite{edm_22_hoogeboom} is able to diffuse and denoise continuous coordinates and discrete attributes simultaneously. The discrete node attributes are encoded into one-hot features and later learned in continuous latent space. In stead of diffusing the graph in continuous space, DiGress \cite{digress_22_vignac} introduces DDPMs to discrete space and diffuses node and edge attributes by applying Markovian transition matrix. The training obejective of categorial attributes in the parameterized reverse kernel is cross-entropy. Depending on specific motion synthesis task, various instructive conditioning variables \cite{motiondiffuse_22_zhang,flame_23_kim,modiff_23_zhao} are injected into denoising kernels. In the field of motion generation, all the diffusion-based methods are DDPMs-based as far as we are aware.

\subsection{Score-Based Generative Models (SGMs)}
The diffusion process perturbs data by a sequence of Gaussian noises with the variance schedule $\sigma_1, \sigma_2, ..., \sigma_T$. The noise distribution at time step $t$ can be represented as $q(x_t | x_0) = \mathcal{N}(x_t; x_0, \sigma_t^2 I)$. Score-based generative models \cite{sgm_19_song,sgm_20_song} intend to restore the original data distribution through estimating the Stein score, which is the gradient of data $\nabla_x log \ p(x)$. A typical technique to learn the score function is score matching \cite{sgm_19_song,interpret_09_lyu}, which trains a score estimation network $s_\theta(x, t)$ towards the direction of the gradient of the log probability density $\nabla_x log \ q(x)$. With $x_t = x_0 + \sigma_t \epsilon$, the training objective takes the form of 
\begin{align}
    & E_{t \sim \mathcal{U}[1, T]}[\lambda(t) \sigma_t^2||s_\theta(x_t, \sigma) - \nabla log \ q(x)||^2] \label{eq:sgm} \\ 
    =& E_{t \sim \mathcal{U}[1, T]}[\lambda(t) \sigma_t^2||-\frac{x_t - x_0}{\sigma_t} - \sigma_t s_\theta(x_t, t)||^2] + const \\
    =& E_{t \sim \mathcal{U}[1, T]}[\lambda(t) ||\epsilon + \sigma_t s_\theta (x_t, t)||^2] + const,
\end{align}
where $\lambda(t)$ represents the weighting function and $\epsilon \sim \mathcal{N}(0, I)$.

Implementations of SGMs in graph generation tasks are relatively limited. In the work of Wu et al.\cite{diffpg_22_wu}, prior information is injected into diffusion bridges. ConfGF \cite{confgf_21_shi} introduces SGMs to conformation generation. A noise conditional score network is desgined to estimate the score from inter-atomic distances and annealed Langevin dynamics is leveraged to generate conformations. DGSM \cite{dgsm_21_luo} learns the gradients of log density score of pair-wise atomic coordinates.

\subsection{Stochastic Differential Equations (Score SDEs)}
By further generalizing discrete time steps to continuous space, stochastic differential equations (Score SDEs) \cite{scoresde_21_song,mlscoresde_21_song} can be used to model the diffusion and denoising process. 

The diffusion process takes the form of stochastic differential equation: 
\begin{equation}
    dx = f(x, t)dt + g(t)dw, \  t \in [0, T],\label{eq:sde}
\end{equation}
where$f(x,t)$, $g(t)$ and $w$ each stands for the drift function, the diffusion function and the standard Wiener process. 

If the diffusion process is defined by Eq.\ref{eq:sde}, the reverse process can be solved by reverse SDE:
\begin{equation}
    dx = [f(x, t) - g(t)^2 \nabla_x log \ q_t(x)] dt + g(t) d \bar{w}, \ t \in [0, T],
\end{equation}
where $\bar{w}$ denotes the standard Wiener process backward in time. With the score function $\nabla_x log \ q_t(x)$, the training objective takes the same form of Eq.\ref{eq:sgm}.

Based on Score SDEs, CDGS \cite{cdgs_22_huang} diffuses molecule graph structures and features through stochastic differential equations and samples generated graphs by solving ordinary differential equations. MOOD \cite{mood_22_lee} builds up stochastic differential equations while incorporating out-of-distribution information. The reverse process is adapted for controlling the diviation from data distribution. During the sampling phase, the gradients of property prediction network are used to encourage futher explorations in the sampling space.

\subsection{Brief Summary of Three Paradigms of Diffusion Models}
DDPMs, SGMs, and Score SDEs differs from each other in many aspacts. 
DDPMs model the forward and reverse process using discrete Markov-chain, which provides strong theoretical guarantees. The forward process of DDPMs includes noise injection and signal attenuation, which are both controlled by the schedule $\beta$. The objective function, KL divergence between 'gronud truth' noisy samples and parameterized ones, provides better guarantee for convergence. 
In SGMs, Langevin dynamics are used to estimate the score of samples. Additionally, SGMs can be relatively sensitive to the choice of noise levels compare to the other two classes of diffusion methods. 
Score SDEs provide a continuous-time stochastic differential equation and regards the generation of sample as the solution of it. Such paradigm is theoretically robust and grounded.

For graph generative models that directly operate on continuous spaces, such as geometric spaces \cite{edm_22_hoogeboom,tordiff_22_jing,motiondiffuse_22_zhang} and latent variable spaces \cite{geoldm_23_xu,hypdiff_24_fu}, they can readily employ the three aforementioned paradigms as their diffusion backbones. However, for those graph generative moethods on discrete spaces, such as graph topologies \cite{digress_22_vignac,edge_23_chen} and discrete feature variables \cite{midi_23_vignac}, they primarily adopt DDPM as the backbone. This is because the forward and reverse process of DDPM can effectively align with their respective model designs.

\subsection{Enhancement of Adaptability of Diffusion on Graphs}

Due to the anisotropic structure of graphs, adapting diffusion to better align the generation of graphs attracts wide attention. 

The most straightforward manner to mitigate the challenge is to introduce diffusion to discrete graph structure. Digress \cite{digress_22_vignac} incorporate transition probability matrices into the discrete diffusion process. Thie conditional probability of noisy graph $G^t = (X^t, E^t)$ takes the form of:
\begin{align}
    q(G^{t} | G^{t-1}) &= (X^{t-1}Q_{X}^{t}, E^{t-1}Q_{E}^{t}), \\
    q(G^t|G) &= (X \bar{Q}_{X}^{t}, E \bar{Q}_{E}^{t}), 
\end{align}
where $\bar{Q}^{t} = Q^1 Q^2 ... Q^t$. By considering distributions of nodes and edges as Bernoulli distributions, EDGE \cite{edge_23_chen} introduces the discrete diffusion that gradually removes or adds edges during the forward or reverse process. Combined with the efficient MPNN \cite{mpnn_17_justin}, EDGE performs well in generating large graphs while preserving statistical properties of original graphs.

By investigating into the distribution of graph data during the forward process, DDM\cite{ddm_23_yang} finds out that adding standard Gaussian noise results in the rapid degradation of signal-to-noise ratios (SNRs), which limits the ability to learn anisotropic structure. Therefore, DDM introduces directional noise, which transform isotropic Gaussian noise into anisotropic noise, to tackle the challenge. Take node faature $X^t$ of noisy graph $G^t = (X^t, E^t)$ for an example, it takes the form of:
\begin{align}
    X^t &= \sqrt{\bar{\alpha}_t} X^0 + \sqrt{1 - \bar{\alpha}_t} \epsilon', \\
    \epsilon' &= \mathrm{sgn}(X^0) \odot |\bar{\epsilon}|, \\
    \bar{\epsilon} &= \mu + \sigma \odot \epsilon, \epsilon \sim \mathcal{N}(0, I).
\end{align}
The customized directional noise ensures the moderate variation of SNRs. To leverage the generation capability of diffusion in concert with the ability of hyperbolic embeddings to capture hierarchical structures, HGDM \cite{hgdm_23_wen} introduces hyperbolic graph variational auto-encoder for hyperbolic node features learning and two score model for generation in both Euclidean and hyperbolic space. HypDiff \cite{hypdiff_24_fu} addresses anisotropy of non-Euclidean structure for graph latent diffusion. Two geometric constraints are introduced to preserve topologic information as well.

In addition to the aforementioned modifications to diffusion, many studies have chosen to fully leverage the excellent modeling capabilities of diffusion on continuous variables. By constructing proper graph encoders and decoders, a number of studies, including NVDiff \cite{nvdiff_22_chen}, Graphusion \cite{graphusion_24_yang}, LDM-3DG \cite{ldm3dg_24_you}, HypDiff \cite{hypdiff_24_fu}, GeoLDM \cite{geoldm_23_xu}, and LatentDiff \cite{latentdiff_24_fu}, perform latent diffusion on implicit representations. In this manner, models exhibit remarkable performances without necessitating complex modifications to diffusion models. 

\subsection{Comparisons and Connections with other generative methods}
Before the rise of diffusion, graph generation was primarily based on VAEs, ARs, GANs, and NFs. 

Compares to VAEs, diffusion models show better capability of generating high quality samples. However, the denoising process makes it slower for diffusion to generate samples.
Compares to ARs, diffusion models can easily generate large samples with high quality. The sequential modeling of ARs leads to relatively weaker ability to model graph with large scale.
By modeling samples with different level of noises, diffusion is more stable to train compares to GANs. NFs provide tractable likelihoods and generate samples in an efficient manner. However, the samples generated by NFs are inferior to those generated by diffusion in terms of both validity and fidelity. The performance difference between E-NF and EDM is the best example of this.

The vast majority of models mentioned in this paper are built on a single diffusion model. However, a small subset of them is constructed on the basis of multiple generative models.

\textbf{With VAEs}:
Many methods, such as Graphusion, GeoLDM, LatentDiff, and LDM-3DG \cite{graphusion_24_yang,geoldm_23_xu,latentdiff_24_fu,ldm3dg_24_you}, performs diffusion and denoising process on the latent space and leverage autoencoders to generate samples givin the prior of latent variables. Such paradigm can be seen as an evolution of tradition VAE-based graph generative methods. By combining the effciency and representational power of latent features with the outstanding generative capabilities of diffusion models, these type of methods exhibit great preformance on graph generation tasks.

\textbf{With ARs:}
Similar to the aforementioned methods combined with VAE, graph diffusion models combined with autoregressive models are also constructed by building the diffusion model on the foundation of the autoregressive model. For example, AutoDiff \cite{autodiff_24_li} and Pard \cite{pard_24_zhao} follow the paradigm that first generate global structures with autoregressive model and then refine local structure based on diffusion models.
This paradigm helps ensure the reasonableness of local structures in diffusion on large graphs and reduces the computational overhead of diffusion.

\begin{table*}[h]
    \centering
    \caption{Summary of diffusion paradigms for graph generation}
    \label{tab:sum_graphdiff}
    \begin{tabular}{ccc}
    \hline
    Paradigm & Tasks & Articles \\
    \hline
    \multirow{3}{*}{DDPMs} & \makecell[c]{AI for Scientific \\ Discovery} & \makecell[l]{EDM \cite{edm_22_hoogeboom}, MDM \cite{mdm_22_huang}, MiDi \cite{midi_23_vignac}, GCDM \cite{gcdm_23_morehead}, DiGress \cite{digress_22_vignac}, GeoDiff \cite{geodiff_22_xu}, DiffBP \cite{diffbp_22_lin}, PMDM \cite{pmdm_23_huang},\\DiffSBDD \cite{diffsbdd_22_schneuing}, TargetDIff \cite{targetdiff_23_guan}, DIFFDOCK \cite{diffdock_22_corso}, DiffAb \cite{diffab_22_luo}, Anand \cite{prostucseq_22_anand}, TSDiff \cite{tsdiff_23_kim}, SILVR \cite{silvr_23_runcie}, \\ SMCDiff \cite{smcdiff_23_trippe}, DiffLinker \cite{difflinker_22_igashov}, PROTSEED \cite{protseed_23_shi}, HierDiff \cite{hierdiff_23_qiang}, D3FG \cite{d3fg_23_lin},GraDe-IF \cite{gradeif_23_yi}, RINGER \cite{ringer_23_grambow},\\GeoLDM \cite{geoldm_23_xu}, DiffMol \cite{diffmol_23_zhang}, GFMDiff \cite{gfmdiff_24_xu}, MUDiff \cite{mudiff_24_hua}, DecompDiff \cite{decompdiff_23_guan}, AbDiffuser \cite{abdiffuser_23_martinkus}, SPDiff \cite{spdiff_24_song},\\LatentDiff \cite{latentdiff_24_fu}, DiffCSP \cite{diffcsp_23_jiao}, DiffCSP++ \cite{diffcsp++_24_jiao}, GemsDiff \cite{gemsdiff_24_klipfel}, CrysDiff \cite{crysdiff_24_song}, GHP-MOFassemble \cite{ghomofassemble_24_park}} \\
    \cline{2-3}
    & Computer Vision & \makecell[l]{MotionDiffuse \cite{motiondiffuse_22_zhang}, Ren et al. \cite{diffmotion_23_ren}, PriorMDM \cite{modiffprior_23_shafir}, Modiff \cite{modiff_23_zhao}, MoFusion \cite{mofusion_22_dabral}, FLAME \cite{flame_23_kim}, \\MLD \cite{mld_23_chen}, Alexanderson \cite{listendenoiseaction_22_alexanderson}, EDGE \cite{edge_22_tseng}, Ahn et al. \cite{canweusediff_23_ahn}, MoDi \cite{modi_22_raab}, BiGraphDiff \cite{bigraphdiff_23_chopin}, MDM \cite{mdm_22_tevet}, \\TCD \cite{tcd_23_saeed},DiffuPose \cite{diffupose_22_choi}, DiffMotion \cite{diffmotion_23_sun}, HumanMAC \cite{humanmac_23_chen}, CommonScenes \cite{commonscenes_23_zhai}, \\DiffuScene \cite{diffuscene_24_tang}, InstructScene \cite{instructscene_24_lin}, Wei et al. \cite{compositional_24_wei}} \\
    \cline{2-3}
    & \makecell[c]{Generic Graph Generation \\ \& Others} & \makecell[l]{DPM-GSP \cite{dpmgsp_23_jang}, DiffSTG \cite{diffstg_23_wen}, HouseDiffusion \cite{housediffusion_23_shabani}, EDGE \cite{edge_23_chen}, EDGE++ \cite{edge++_23_chen}, Lee et al. \cite{microstructure_23_lee},\\NAP \cite{nap_23_lei}, SaGess \cite{sagess_23_limnios}, DIFUSCO \cite{difusco_23_sun}, DDM \cite{ddm_23_yang}, HypDiff \cite{hypdiff_24_fu}, Bergmeister et al. \cite{efficient_24_bergmeister}, Pard \cite{pard_24_zhao}} \\
    \hline
    \multirow{2}{*}{SGMs} & \makecell[c]{AI for Scientific \\ Discovery} & \makecell[l]{Wu et al. \cite{diffpg_22_wu}, ConfGF \cite{confgf_21_shi}, ColfNet \cite{colfnet_22_du}, DGSM \cite{dgsm_21_luo}, ProteinSGM \cite{proteinsgm_22_lee},\\ Arts et al. \cite{dff_23_arts}, VoxMol \cite{voxmol_23_pinheiro}} \\
    \cline{2-3}
    & \makecell[c]{Generic Graph Generation \\ \& Others} & \makecell[l]{EDP-GNN \cite{edpgnn_20_niu}, DruM \cite{drum_23_jo}, SLD \cite{sld_23_yang}} \\
    \hline
    \multirow{3}{*}{Score SDEs} & \makecell[c]{AI for Scientific \\ Discovery} & \makecell[l]{CDGS \cite{cdgs_22_huang}, DiffMD \cite{diffmd_23_wu}, EigenFold \cite{eigenfold_23_jing}, JODO \cite{jodo_23_huang}, NeuralPLexer \cite{neuralplexer_23_qiao}, EEGSDE \cite{eegsde_23_bao}, SD \cite{sd_24_hsu}, \\MuDM \cite{mudm_24_han}, DiffBindFR \cite{diffbindfr_24_zhu}} \\
    \cline{2-3}
    & Computer Vision & \makecell[l]{DiffuseSG \cite{diffusionsg_24_xu}} \\
    \cline{2-3}
    & \makecell[c]{Generic Graph Generation \\ \& Others} & \makecell[l]{GraphGDP \cite{graphgdp_22_huang}, GSDM \cite{gsdm_22_luo}, NVDiff \cite{nvdiff_22_chen}, HGDM \cite{hgdm_23_wen}, DiffusionNAG \cite{diffusionnag_23_an}, \\Diff-POI \cite{diffpoi_23_qin}, Graphusion \cite{graphusion_24_yang}} \\
    \hline
    \end{tabular}
\end{table*}

\section{Applications}
Diffusion models not only achieved outstanding performance for images and language data, but also show great potential in the field of special data structure generation. In this section, we will elaborate on applications of diffusion on AI for scientific discovery, computer vision, generic graph generation and other forms of graph generative tasks.

\subsection{AI for Scientific Discovery}
\begin{figure*}[t]
    \centering
    \includegraphics[width=0.8\linewidth]{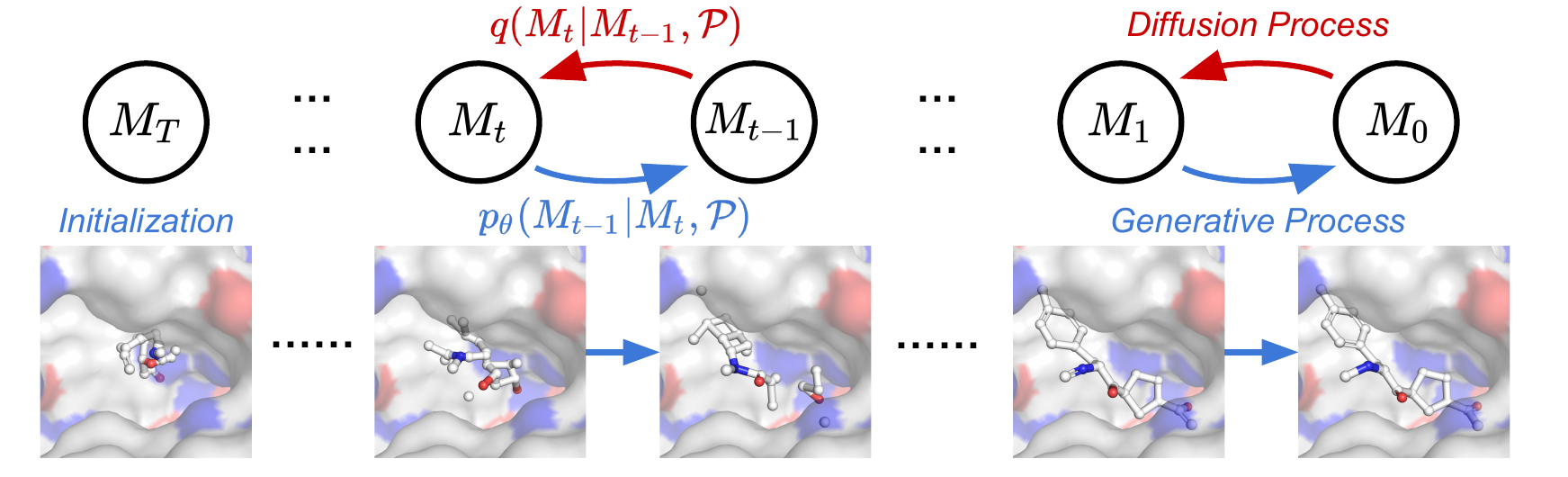}
    \caption{An illustration of diffusion-based molecule generation \cite{targetdiff_23_guan}}
    \label{fig:targetdiff}
\end{figure*}

Molecule generation tasks aim at generating valid molcules with desirable  properties. Depending on the size and nature of molecules and the given information, the generation tasks can be further categorized into sub-tasks such as {\itshape de novo} molecule design, conformation design, {\itshape de novo} ligand design, ligand docking, and protein design. A typical paradigm for diffusion-based molecule generation is shown in the Figure~\ref{fig:targetdiff}.

\subsubsection{{\itshape De novo} molecule design}
{\itshape De novo} molecule design is suppoesed to generate novel, unique, and valid molecule graphs or geometries from scratch. Initially, early attempts built on variational autoencoders (VAEs) and generative adversarial networks (GANs) focus on 2D molecule graphs, whereas recent research shifts attention towards 3D molecule geometries. However, due to the discrete nature of graphs, how to diffuse and sample discrete features and adjacency matrix still remains a major challenge. 

\textbf{2D Molecule Desgin.}
DiGress \cite{digress_22_vignac} is among the few diffusion-based methods that specifically generate 2D molecule graphs. It diffuses graph data with graph edits and learns the reverse process using a graph transformer network. DiGress shows comparable performance to other one-shot generative methods in terms of diversity, novelty and other evaluation metrics.
CDGS \cite{cdgs_22_huang} introduces a hybrid graph noise prediction approach for global and local context extraction together with a ODE solver that promotes efficient graph sampling. 

\textbf{3D Molecule Design.}
The first diffusion-based method for {\itshape de novo} 3D molecule design is EDM \cite{edm_22_hoogeboom}. It operates on the continuous Cartesian coordinates and discrete atomic number in the denoising process. The design of transition kernel follows EGNN \cite{egnn_21_satorras}, which is an E(n) equivariant graph neural network. EDM shows superior performance compared to previous methods in generating medium size molecules.
MDM \cite{mdm_22_huang} introduces dual equivariant score neural networks, leveraging two SchNets \cite{schnet_17_schutt} to capture global and local inter-atomic relationships, respectively. Additionally, latent variables are integrated into the diffusion and reverse process to encourage full exploration. Experimental results show that MDM performs better in large-size molecule generation.
The model built by Wu et al. \cite{diffpg_22_wu} employs SGM instead of DDPM. The author propose a handful of energy function for integrating physical and statistical prior knowledge.
GCDM \cite{gcdm_23_morehead} achieves geometry-completion through GCPNET, a graph neural network with similar geometry learning philosophy as ColfNet \cite{colfnet_22_du}.
JODO \cite{jodo_23_huang} is able to jointly generate 2D molecule graphs and 3D conformations.
DiffLinker \cite{difflinker_22_igashov} introduces diffusion models to molecular linker design. The architecture of denoising kernel is similar to EDM \cite{edm_22_hoogeboom}.
SILVR \cite{silvr_23_runcie} is conditioned on EDM \cite{edm_22_hoogeboom} for fragement merging and linker generation. The model is able to explore new chemical space with ideal conditions cheaply.
Research by Arts et al. \cite{dff_23_arts} leverage SGMs and force field to learn coarse-grained molecular dynamics.
HierDiff \cite{hierdiff_23_qiang} first generates fragment representations instead of deterministic fragment, and then construct atom-level 3D structures. 
Unlike other studies that perform diffusion on feature and position space, GeoLDM \cite{geoldm_23_xu} and LDM-3DG \cite{ldm3dg_24_you} map molecule geometries into latent space and performs diffusion process in latent space.
GFMDiff \cite{gfmdiff_24_xu} addresses the challenge of bond generation in 3D molecule generation and builds GFLoss based on pre-defined rules. Such loss term guides the molecule in generating authentic bonds at each sampling step.
Instead of point clouds, VoxMol \cite{voxmol_23_pinheiro} voxelizes molecules in voxel grid, where the length of each cubic grid is .25\AA. Such approach does not requires to know the number of atoms that need to be generated and scales better compares to previous methods. 

\textbf{Joint 2D \& 3D Molecule Design.}
Most methods for 3D molecule design operates on 3D point clouds and generate bonds after the sampling phase. Technically speaking, such paradigm is much easier since it does not explicitly generate bonds during the diffusion process. 
MiDi \cite{midi_23_vignac} designs the reverse kernel named rEGNNs and directly predicts the existence of bonds rather than using predetermined rules. This allows for the simultaneous generation of 2D molecule graphs and their corresponding conformers, resulting in enhanced stability of bonds and validity of molecules. 
MUDiff \cite{mudiff_24_hua} introduces the transformer-based MUformer, which encodes structural, spectural, and velocity features. Guassian noises and discrete noises are separately added to atom features and edge features.

\textbf{Conditional Molecule Design.}
Though most methods mentioned above are designed with the native ability to generate molecule with desired properties, some studies are dedicated to enhancing such capabilities. For those methods that are not specifically designed for this task, the common approach to perform property conditioned molecule generation takes two steps. The first step includes training an graph neural network that predicts certain molecule properties, and the second step is to concatenate desired properties with node features at both training and sampling stages. 
EEGSDE \cite{eegsde_23_bao} incorporates an energy guidance function into the stochastic differential equation to guide denoising process.
MuDM \cite{mudm_24_han} is a training-free methods, which leverages time-independent posterior approximation and MC sampling. It also achieves remarkable performance in multi-condition generation task by introducing multi-condition guidance derived from probabilistic graphs.

\subsubsection{Conformation generation and optimization}
The precise yet costly molecular dynamics (MD) is crucial for atomistic system simulation. Recent studies predicts 3D molecule conformations with the assistence of deep generative methods with less computational cost.
ConfGF \cite{confgf_21_shi} is one of the first few diffusion-based methods in this area. It proposes to learn gradient fields of atomic coordinates' log density and regard them as pseudo-forces that act on atoms. With the gradient fields, samples are generated using Langevin dynamics. 
DGSM \cite{dgsm_21_luo} argues that previous methods disregard the long-range interactions between non-bonded atoms. During the denoising process, representations and coordinates in the dynamic graph are estimated using MPNN \cite{mpnn_17_justin}. 
GeoDiff \cite{geodiff_22_xu} lays stress on discussing the induced equivariance of Markov chains evolving with equivariant kernels in diffusion process. An equivariant convolutional layer named GFN is designed to be the transition kernel for molecule graph learning. 
ColfNet \cite{colfnet_22_du} comes up with an orthonormal-based local frame for geometries extraction. 
Unlike other methods that inject noise to coordinates, Torsion Diffusion \cite{tordiff_22_jing} predicts conformers while only acting on torsion angles, which drastically reduces the sample space. Therefore, an extrinsic-to-intrinsic score model is introduced to convert 3D coordinates to scores in torsional space. Finally, a Boltzmann generator is specifically designed for the chemical system without costly molecular dynamics. 
DiffMD \cite{diffmd_23_wu} introduces ScoreSDEs to molecular dynamics along wih a equivariant geometric transformer network. 
For conformation generation of macrocycle backbones, RINGER \cite{ringer_23_grambow} is able to effectvely encode ting geometry and handle the cyclic nature of macrocycles and side chains.
DiSCO \cite{disco_24_lee} employs SE(3) Schr\"{o}dinger bridge to predict atom cordinates and aligns distribution of samples with ground-truth energy landscape.

\subsubsection{{\itshape De novo} ligand design}
Instead of generating molecules from scratch, ligand design involves designing molecules specifically bind to target receptors. Structure-based drug design (SBDD) aims at predicting the optimal position and oriantation of a small molecule ligand with high affinity and specificity that bind to specific proteins. 
In the research of DiffBP \cite{diffbp_22_lin}, the researcher mentions that previous research \cite{graphbp_22_liu, arligand_21_luo, pocket2mol_22_peng} based on auto-regressive models generate ligand in a sequential manner, which leads to several dilemmas. The sequential modeling of molecules is inconsistent to the global interactions of atoms and may lead to early stopping. DiffBP proposes a target-aware diffusion process to model and generate atoms in molecules all at once. Though it surpasses previous methods at some level, the model still suffers from validity of molecules and the drug-likeness of sub-structures. 
DiffSBDD \cite{diffsbdd_22_schneuing} also performs one-shot molecule generation based on DDPMs and introduces two strategy, protein-conditioned generation and ligand-inpainting generation. Along with the proposed innovation points is a determined binding dataset derived from Binding MOAD \cite{bindingmoad_05_hu}. 
TargetDiff \cite{targetdiff_23_guan} designs a simple SE(3) GNN kernel to connect generative models and binding affinity ranking to evaluate generated samples. Therefore, the model is capable to improve binding affinity prediction in an unsupervised manner. 
PMDM \cite{pmdm_23_huang} leverages the dual equivariant score kernel in MDM \cite{mdm_22_huang} for pocket-based ligand design and achieves outstanding performance in multiple metrices. 
D3FG \cite{d3fg_23_lin} decompose molecules into different components and is capable to generate molecules with realistic structures, drug properties, and binding affinity.
AutoDiff \cite{autodiff_24_li} introduces a diffusion-based autoregressive model, which generate molecules motif-by-motif. Diffusion model is adopted to predict torsional angles between fragments.
DecompDiff \cite{decompdiff_23_guan} also address SBDD task by treating atoms in ligand molecules differently. By decomposing molecules into scaffolds and arms and introducing validity guidance, it generates ligand molecules with better drug-likeness and synthesizability.

\subsubsection{Ligand docking}
Apart from generating novel ligand, molecular ligand docking plays a essential role in drug discovery by predicting positions and conformations of given molecules bound to target proteins, is also a crucial task in drug discovery. 
Unlike previous methods based on regression frameworks, DIFFDOCK \cite{diffdock_22_corso} designs a diffusion model over degrees of freedom, including translational, rotational, and torsional. The implementation of diffusion-based paradigm significantly lifts the success rate of predictions and reduces inference time. 
EDM-Dock \cite{edmdock_23_masters} utilizes two EGNNs for extraction of geometric and chemical information of proteins and ligands respectively and a multilayer perceptron for protein-ligand pairwise distance predictions. 
DPL \cite{dpl_22_Nakata} introduce a novel reverse kernel for their diffusion framework, which consists of input featurization, residual feature update, and equivariant denoising. 
NeuralPLexer \cite{neuralplexer_23_qiao} is a framework capable of generating protein-ligand complex. The paper proposes contact prediction module to generate inter-molecular distance distributions, contact maps, and pair representations.
DiffBindFR \cite{diffbindfr_24_zhu} emphasizes the importance of protein side chian conformation modeling and proposes full-atom graph construction. Pocket side chan torsional score, ligand rotatable bond torsional score, ligand rotational score, and ligand transitional score are introduced to solve the reverse SDE.

\subsubsection{Protein design}
In the field of protein design, there are three main tasks, which are antigen-specific antibody design, protein sequence and structure design, and motif-scaffold design. 

\textbf{Antibody design:}
Antibody design aims to generate antibodies bind to target antigen. 
Focusing on this particular task, DiffAb \cite{diffab_22_luo} generates sequences and structures of complementarity-determining regions (CDR) of antibodies conditioned on antigen structures. Besides, the model is capable to design side-chain orientations of amino acids.
AbDiffuser \cite{abdiffuser_23_martinkus} introduce a novel SE(3) network, which implicitly models residue-to-residue relations and handles sequences with various length better. 

\textbf{Protein sequence and structure design:}
Protein sequences and structure design involves creating and modifying protein sequences and structures to improve stabilities or other properties. 
Research by anand \cite{prostucseq_22_anand} introduces diffusion-based method to generate protein structures, sequences and rotamers successively. However, this approach may cause inconsistencies and expensive inferences. 
To tackle this issue, PROTSEED \cite{protseed_23_shi} proposes to co-design structures and sequences. The model first extract constraints from a trigonometry-aware context encoder. An equivariant decoder is also designed to update positions of $C_\alpha$, rotation matrix, and amino acid types. 
Based on Score-SDE, ProteinSGM \cite{proteinsgm_22_lee} generates 6D coordinates of proteins. 
LatentDiff \cite{latentdiff_24_fu} first maps 3D protein structures to latent spaces using autoencoders and integrates latent diffusion to protein structure genration.

\textbf{Motif-scaffold design:}
Motif-scaffold design scaffoles for desired motifs, a kind of sub-structures that play important roles in the function of proteins. SMCDiff \cite{smcdiff_23_trippe} is proposed to design stable scaffold to support a desired motif. Based on EGNN \cite{egnn_21_satorras}, ProtDiff generates protein backbone in the first place, and SMCDiff later generates scaffolds. For inverse protein folding, GraDe-IF \cite{gradeif_23_yi} predicts joint distribution of animo acids conditioned on node properties. EigenFold \cite{eigenfold_23_jing} recognize proteins as dynamic structural with conformational flexibility. The model introduces harmonic diffusion and view the reverse process as a cascading-resolution process.

\subsubsection{Other applications}
Apart from applications on drug discovery, many studies proposes diffusion-based graph generative methods on many other applications on scientific discovery.

\textbf{Crystal structure prediction:}
The goal of crystal structure prediction is to generate stable 3D structures based on compositions of compounds. Traditional methods mainly integrates Density Functional Theory (DFT) \cite{dft_65_kohn} to compute the ennergy of compound structures and guide the search to local minima of energy surface. Applications of deep generative methods on this taks could reduce the computational cost. 
DiffCSP \cite{diffcsp_23_jiao} is one of the first methods that focus on this task. It jointly predicts atom fractional coordinates and lattice vectors. 
On this basis, DiffCSP++ \cite{diffcsp++_24_jiao} emoloys space group constraint, which is presented in the form of basis constraint of O(3)-invariant logarithmic space of lattice matrix and the Wyckoff position costraint of atom coordinates. 
GemsDiff \cite{gemsdiff_24_klipfel} introduce a new metric named Frechet ALIGNN Distance (FAD) to provide robust evaluation of generated crystals.
CrysDiff \cite{crysdiff_24_song} design a reconstruction pre-training task and enhance performance on crystal property prediction, which is constrained by limited number of labeled data in previous studies.
GHP-MOFassemble \cite{ghomofassemble_24_park} presents a method for metal-organic frameworks (MOFs) discovery, which shows great potential in $\mathrm{CO}_2$ capture.
Lee et al. \cite{microstructure_23_lee} utilizes implicit diffusion moedels for microstructure characterization and reconstruction. 

\textbf{Others:}
Brain Diffuser \cite{braindiffuser_23_chen} shapes the structural brain networks from tensor images and shows better ability in exploiting structutal connectivity features and disease-related information. 
By adapting the problem of crowd simulation to spatial graph generation, SPDiff \cite{spdiff_24_song} designs multi-frame rollout training algorithm to simulate pedestrian trajectories.

\subsection{Computer Vision}
When it comes to the applications of generative graph diffusion models in the field of computer vision (CV), motion generation and scene graph generation are the two most widely researched directions.

\begin{figure*}[t]
    \centering
    \includegraphics[width=0.8\linewidth]{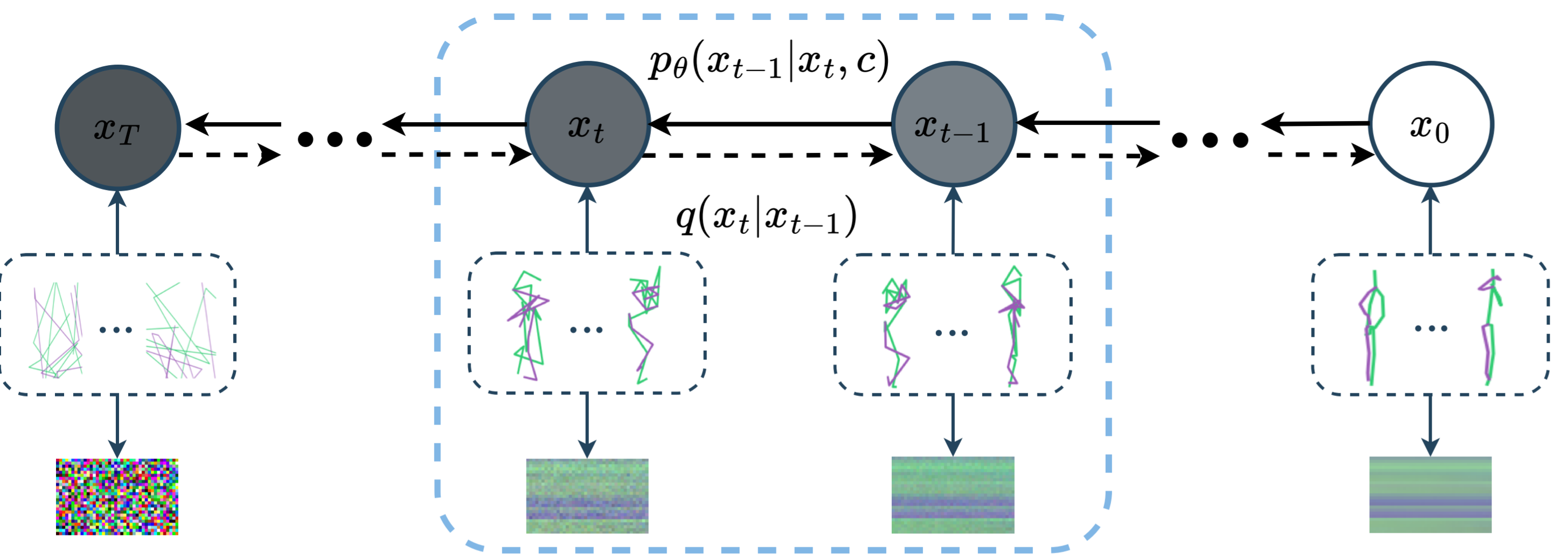}
    \caption{An illustration of diffusion-based motion generation \cite{modiff_23_zhao}}
    \label{fig:modiff}
\end{figure*}

Human motion generation plays an indispensable role in virtual character animation. In essence, it involves creating virtual characters and controlling their movement in a way that is believable, natural, and congruent with human motion patterns. Research in this area primarily regards human poses in formations like 3D coordinates of joints or other variations. Human poses are highly constrained due to their relatively fixed nature, which contrasts with the inherent complexity of molecule geometries. Therefore the inconvenience of edge generation could be avoided. To capture the nuances of human motions, some research model human skeletons as 3D point clouds, which solely takes 3D coordinates of joints into account. However, more studies incorporate joints' angular and linear velocities, and roations into their models to represent the complexity and subtleties of human motions and create immersive poses. A simple illustration of diffusion-based motion generation methods is shown in Figure~\ref{fig:modiff}.

\subsubsection{Human motion synthesis}
Human motion synthesis is prone to generate human motion sequences under the instructions of given inputs such as texts, audio, and actions. MotionDiffuse \cite{motiondiffuse_22_zhang} is the first diffusion-based motion generation framework in this field, which is able to generate diverse human motions. In order to achieve fine-grained manipulation on body parts, part-aware text controlling is added to instruct each body part with different texts. Another text-conditioned signal named time-varied controlling is designed to synthesize long motion sequences with multiple actions. In a similar vein, the method proposed by Ren et al. \cite{diffmotion_23_ren} fuses BERT embeddings of texts into time embeddings for text-driven motion generation. FLAME \cite{flame_23_kim} is also pioneering in diffusion-based motion generation. Time-step and motion-length tokens are feed into the transformer kernel to make the model time-aware and length-aware. 

Using U-Net\cite{unet_15_ronneberger} as the backbone of denoising kernel, Modiff \cite{modiff_23_zhao} implements DDPMs for action-conditioned motion generation. MoFusion \cite{mofusion_22_dabral} tackles motion synthesis though similar ideas with a time-varying weight schedule to achieve temporally plausibility and semantically accuracy. Unlike other methods that use U-Net \cite{unet_15_ronneberger} backbone, MDM \cite{mdm_22_tevet} employs a transformer-based encoder and geometric losses with dedicated terms to promote generation. PriorMDM \cite{modiffprior_23_shafir} leverages MDM \cite{mdm_22_tevet} as a generative prior and significantly improve qualities of samples in multiple new genration tasks using limited examples. For long sequences generation, DoubleTake, which generate long motions and parallel solutions, is proposed. MLD \cite{mld_23_chen} designs the transformer-based autoencoder with long skip connections to map motion sequences into latent space. This allows the model to produce vivid motion while reduing the computational cost.

In the domain of audio-driven motion synthesis, Alexanderson et al. \cite{listendenoiseaction_22_alexanderson} leverages Conformers \cite{conformer_20_gulati} for motion generation and classifier-free guidance for style control. EDGE \cite{edge_22_tseng} combines Jukebox \cite{jukebox_20_dhariwal} and transformer-based diffusion model to dance generation. By adding the contact consistency loss and physical foot contact score, the model is able to eliminate physical implausibility without directly modelling the physical skeletons. SceneDiffuser \cite{scenediffuser_23_huang} proposed a unified framework for generating poses, motions, dexterous grasps, path and motion planning conditioned on 3D scenes. 

Rather than transforming motion sequences into pseudo images, MoDi \cite{modi_22_raab} leverages structure-aware neural filters and 3D convolutions which promotes dedicated control on each joint. For two-person generation, ComMDM is introduced to coordinate two separate MDMs \cite{mdm_22_tevet} and infuse two motions. BiGraphDiff \cite{bigraphdiff_23_chopin} is dedicated to generate interaction sequences of two individuals based on given text. The denoising kernel separately embeds joints of each person and jointly predicts the output using the bipartite graph struture. DiffuPose \cite{diffupose_22_choi} adopts graph convolutions instead of transformer-based network as the denoising kernel. The model is also employed to 2D-to-3D lifting human pose estimation.

\subsubsection{Human motion prediction}
Rather than generating motions from other modalities of information, human motion prediction aims at forecasting motions based on given observed sequences. HumanMAC \cite{humanmac_23_chen} addresses the motion prediction problem in a masked completion fashion. In order to encode and decode motion sequence while preserving both current and periodic properties, Discrete Cosine Transform (DCT) and inversed Discrete Cosine Transform (iDCT) are proposed. Another study on diffusion-based motion prediction by Ahn et. al. \cite{canweusediff_23_ahn} proposes two version of Transformer-based motion denoising kernel, which firstly process spatial and temporal information and then combine them in either serial or parallel manner. TCD \cite{tcd_23_saeed} regards both observations and predictions as a single sequence with noises and is capable to handle long-term forecasting horizon. DiffMotion by sun et al. \cite{diffmotion_23_sun} performs motion reconstruction and refinement separately with a transformer-based network and a multi-stage graph convolutional network.

\subsubsection{Scene Graph Generation}
Scene graph generation aim at providing controllable scene synthesis for industrial applications. Specifically, objects are represented by nodes while relationships among them are represented by edges.

Different from previous retrieval-based and semi-generative methods, CommonScenes \cite{commonscenes_23_zhai} introduces fully generative model for scene synthesis. It shows the capablity to jointly capture global inter-object relationships and local shape cues.
DiffuScene \cite{diffuscene_24_tang} introduces shape latent diffusion for shape retrieval, which performs diffusion process on object attributes and geometries. 
DiffuseSG \cite{diffusionsg_24_xu} proposes joint scene graph imange generation, a novel task in this field. Rather than simply models object attributes, it jointly samples adjacency matrix, node attributes, bounding boxes, and edge attributes. 
Also focusing on joint scene graph image generation, Wei et al. \cite{compositional_24_wei} introduces LLM to facilitate better understanding on graphs and inter-object relationships.
To address the fact that methods and data within this field being primarily focused on single scenes, resulting in a lack of flexibility and controllability,  InstructScene \cite{instructscene_24_lin} integrates semantic graph prior and layout decoders to overcome these limitations. It shows comparable performance in multiple downstream tasks in a zero-shot manner.



\subsection{Generic Graph Generation}
In this sub-section, we aim to provide a comprehensive overview on methods that primarily on Generic graph generation. This methods are designed to better adapt diffusion methods to graph generative task without domain restrictions.

EDGE \cite{edge_23_chen} introduces diffusion models to large graph generation. The diffusion process involves gradually remove edges until the graph is empty. The model also manage to avoid the generation of excessive edges by focusing on a small portion of nodes.
Additionally, EDGE++ \cite{edge++_23_chen} proposes a degree-specific noise schedule to fully utilize the computational power of diffusion model. The volume-preserved sampling strategy is also introduced to mitigate the inconsistency between behavior of active nodes in diffusion and denoising process.
SaGess \cite{sagess_23_limnios} augments DiGress \cite{digress_22_vignac} with a generalized divide-and-conquer framework. 
For 2D graph generation, EDP-GNN \cite{edpgnn_20_niu} is the pioneer in incorporating Score SDEs to discrete graph adjacency matrix generation. 
Based on this, GraphGDP \cite{graphgdp_22_huang} proposes the denoising kernel named position-enhanced graph score network to make further use of positional information. 
Instead of sampling only on edges attributes, GSDM \cite{gsdm_22_luo} leverages Score SDEs on node representations and graph spectrum space to generate graphs with great efficiency, which is shown in results on generic datasets and molecule datasets. 
Inspired by VAEs, NVDiff \cite{nvdiff_22_chen} takes the VGAE structure and by first samples latent representations of graphs and then decodes them into representations node and edge. 
SLD \cite{sld_23_yang} takes graph generation task in a similar manner. 
GraphARM \cite{ardiff_23_kong} combines autoregressive models with diffusion models by introducing a node-absorbing diffusion process. 
Inspired by latent diffusion \cite{stablediffusion_22_rombach}, Graphusion \cite{graphusion_24_yang} encodes graph into topology-injected latent space, and guides the denoising process with latent self-guidance, which predicts latent variables with pseudo-labels from K-means clustering. 
In order to preserve invariance, LDM-3DG \cite{ldm3dg_24_you} leverages a pretrained autoencoder to map graph into low-dimension latent space in a data-driven manner.
DruM \cite{drum_23_jo} proposes to directly predict the the destination of the generative process using a mixture of endpoint-conditioned process. 
Specifically designed for hyperbolic graph, HGDM \cite{hgdm_23_wen} shows promising performance in extracting complex geometric features of hyperbolic embeddings. 
HypDiff \cite{hypdiff_24_fu} also addresses anisotropy of non-Euclidean structure for graph latent diffusion by introducing two geometric constraints.
To generate graph with large scale efficiently, Bergmeister et al. \cite{efficient_24_bergmeister} proposes a method that first construct global structure then refines local details. 

In order to make graph-structure prediction in supervised learning, DPM-GSP \cite{dpmgsp_23_jang} performs diffusing and denoising process on targets of node classification tasks and achieves outstanding performance. 
Spatio-temporal graph (STG) forecasting is the problem that requires accurate predictions of signals generated by graphs and their historical observations. 
DiffSTG \cite{diffstg_23_wen} is the first research that combines DDPMs with spatio-temporal graph neural networks UGNet. 
The proposed framework is capable to capture temporal dependencies and spatial correlations. 
DDM \cite{ddm_23_yang} addressses the anisotropic structures in graph data and proposes a novel diffusion model by introducing a data-dependent noise schedule. 
DiffusionNAG \cite{diffusionnag_23_an} is a novel transferable task-guided framework for neural architecture generation. 
The model adoptes gradients of the surrogate model to facilitate the model to generate ideal results for given tasks.

In the field of knowledge graph embedding (KGE), KGDM \cite{kgdm_24_long} transforms the problem of entity prediction into conditional entity generation. On this basis, FDM \cite{fdm_24_long} argues that patterns among entities and ralations of fact come from low-dimensional fact space and introduces a learnable fact embedding model to bridge the gap between diffusion and discrete knowledge graph. 

\subsection{Other Applications}
Apart from graph generation applied to the research fields mentioned above, there are also several interesting studies on other applications. 

\begin{figure*}[htbp]
  \centering
  \begin{minipage}[t]{0.9\textwidth}
  \centering
  \includegraphics[width=14cm]{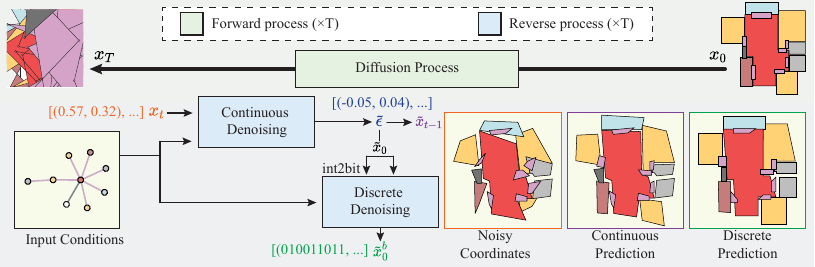}
  \end{minipage}
  \begin{minipage}[t]{0.9\textwidth}
  \centering
  \includegraphics[width=14cm]{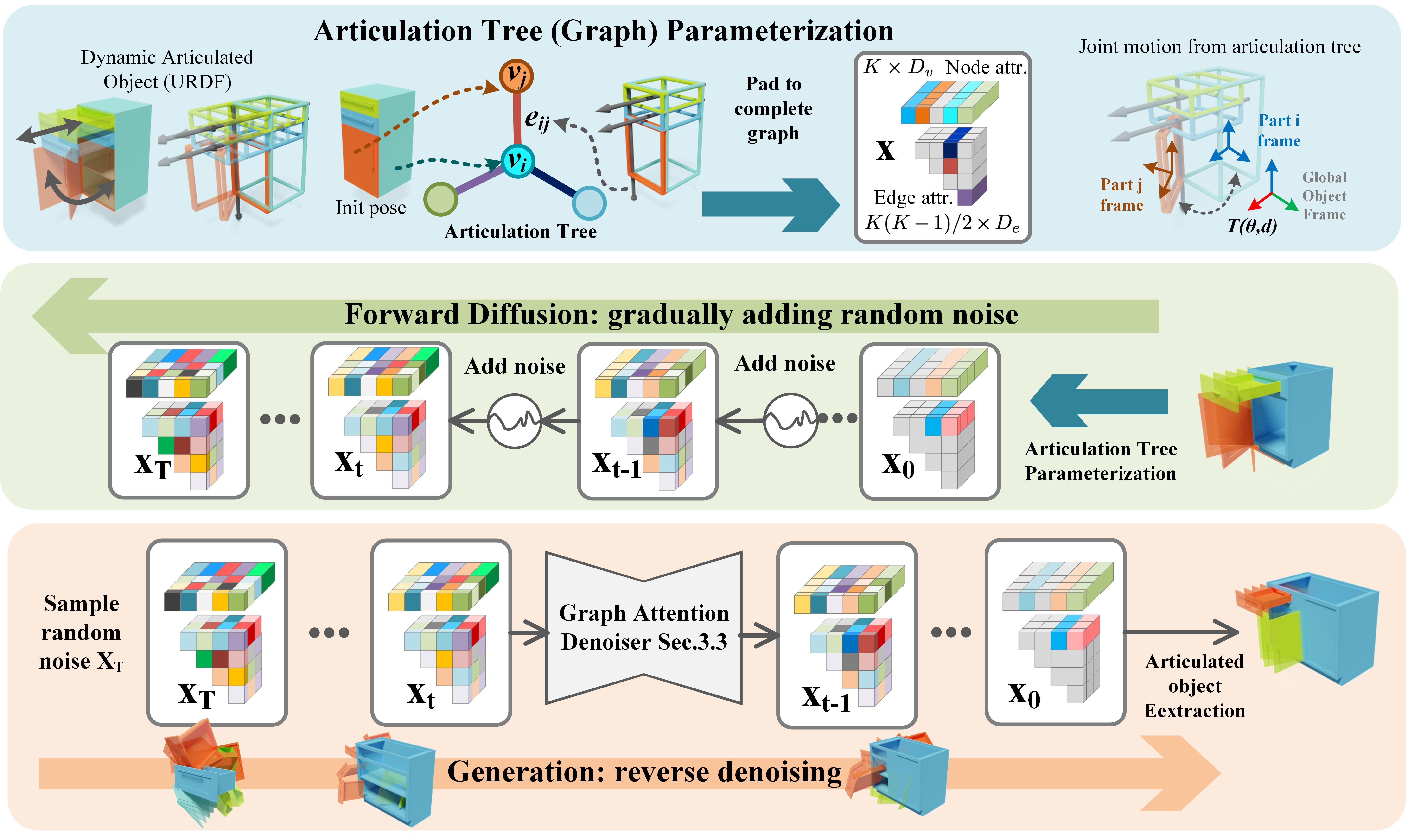}
  \end{minipage}
  \caption{Illustrations of HouseDiffusion\cite{housediffusion_23_shabani} and NAP\cite{nap_23_lei}. The upper one is the overall structure of HouseDiffusion while the lower one is the overview of NAP.}
  \label{fig:housediffusionnap}
\end{figure*}

In order to solve NP-complete (NPC) problems, which are fundamental challenges in combinatorial opimization problems, DIFUSCO \cite{difusco_23_sun} employs graph-based diffusion models to generate high-quality resolutions. 
Continuous Gaussian noise and discrete Bernoulli noise are tested and the latter one exhibits better accuracy and scalability. 
Diff-POI \cite{diffpoi_23_qin} samples user spatial preference for next point-of-interest recommendations. 
DiffKG \cite{diffkg_24_jiang} integrates diffusion with knowledge graph learning in knowledge-aware recommendations.
Inspired by nonequilibrium thermodynamics, Lu et al. \cite{modeling_23_lu} introduces analog and discrete diffusion with neighbour reprentations to {\itshape de novo} bio-inspired spider web mimics.
HouseDiffusion \cite{housediffusion_23_shabani} introduces diffusion to vector floorplan generation. Given input graphs where each vertice or edge represents a room or a door, the model samples polygonal loops of vertices and edges. Each loop is composed of a set of sequence of 2D coordinates of corners. Results indicate that HouseDiffusion is able to generate floorplan with great precision. 
NAP \cite{nap_23_lei} applies DDPM to 3D articulated object synthesis. Articulation tree parameterization is introduced map dynamic articulated objects to articulated graph. By sampling on node and edge attributes, 3D articulated objects are geneated. Illustrations of HouseDiffusion and NAP are shown in Figure~\ref{fig:housediffusionnap}.



\section{Metrics and datasets}
In this section, we will elaborate on evaluation metrics and datasets adopted in the above applications.

\subsection{Evalutaion metrics}
\subsubsection{Metrics for AI for scientific discovery tasks}
We classify evaluation metrics based on different molecule generation tasks: de novo molecule design, Conformation design, de novo ligand design, ligand docking, and protein generation. 

(1) De novo molecule design

Versatile evaluation metrics for 2D and 3D de novo molecular design are \textbf{Validity}, \textbf{Uniqueness}, \textbf{Novelty}, and \textbf{Quality}. These metrics are mostly calculated based on generated smiles strings instead of 3D geometries. 

\begin{itemize}
\item[$\bullet$] Validity: the number of valid molecules compares to generated molecules. Whether molecules are valid is generally determined by compliance with the valence rules in RDkit.
\item[$\bullet$] Uniqueness: the number of unique molecules compares to generated molecules. This reflects the models' abilities to design non-repetitive molecules.
\item[$\bullet$] Novelty: the number of new molecules among all generated molecules. A molecule that makes no appearance in the training set is considered as new molecule.
\item[$\bullet$] Quality: the proportion of molecules with good-quality filtered by certain chemical rules.
\end{itemize}

\begin{equation}
  \text{Validity} = \frac{\# \text{valid \ molecules}}{\# \text{generated \ molecules}},
\end{equation}

\begin{equation}
  \text{Uniqueness} = \frac{\# \text{unique \ molecules}}{\# \text{generated \ molecules}},
\end{equation}

\begin{equation}
  \text{Novelty} = \frac{\# \text{novel \ molecules}}{\# \text{generated \ molecules}},
\end{equation}

\begin{equation}
  \text{Quality} =\frac{\# \text{good-quality \ molecules}}{\# \text{generated \ molecules}},
\end{equation}

Apart from the basic metrics listed above, there are some metrics raised or adopted by several studies. Stability of generated molecules is usually measured by \textbf{Atom stability} and \textbf{Molecule stability}. MiDi \cite{midi_23_vignac} introduce \textbf{atom total variation (AtomTV)} and \textbf{bond total variation (BondTV)}. 
\begin{itemize}
\item[$\bullet$] Atom stability: the number of atoms with the correct valence state compares to the total number of atoms.
\item[$\bullet$] Molecule stability: the percentage of generated molecules whose atoms are stable altogether.
\item[$\bullet$] AtomTV: the distance of distribution of atom types between generation and test set.
\item[$\bullet$] BondTV: the distance of distribution of bond types between generation and test set.
\end{itemize}

(2) Conformation design

Metrics for this task involve measuring the quality and diversity of generated conformations. By calculating the \textbf{root mean square deviation (RMSD)} of heavy atoms between generated samples from references, \textbf{coverage scores (COV)} and \textbf{matching scores (MAT)} are the most commonly adopted for performance evaluation.

In order to get RMSD, which measures the difference between two conformations, the reference conformation $R^*$ are firstly aligned by translation and rotation to the generated conformation $R$. Let conformation $\hat{R}$ be the reference conformation after alignment and the RMSD can be represented by: 
\begin{equation}
  \operatorname{RMSD}(\mathbf{R}, \hat{\mathbf{R}})=\left(\frac{1}{n} \sum_{i=1}^n\left\| \mathbf{R}_i - \hat{\mathbf{R}}_i \right\|^2\right)^{\frac{1}{2}},
\end{equation}
where $n$ is the number of heavy atoms. Built upon RMSD, COV and MAT each measures the diversity and accuracy of generated conformations, respectively.

\begin{align}
  & \operatorname{COV}(S_g, S_r) \notag \\
  &= \frac{1}{|S_r|} | \{ \mathbf{R} \in S_r \mid \operatorname{RMSD}(\mathbf{R, \hat{\mathbf{R}}}) < \delta, \ \hat{\mathbf{R}} \in S_g \}|, \\ 
  & \operatorname{MAT}(S_g, S_r) = \frac{1}{|S_r|} \sum_{\mathbf{R} \in S_r} \min_{\hat{\mathbf{R}} \in S_g} \operatorname{RMSD}(\mathbf{R}, \hat{\mathbf{R}}),
\end{align}
where $S_g$ and $S_r$ each stands for the generated and reference conformation set and $\delta$ is the predefined threshold. Formally, COV evaluates the percentage of reference conformations got covered by conformations in the generated set, while MAT shows the average RMSD score of between conformations in reference set and their closest generated sample.

(3) De novo ligand design

Compare to de novo molecule design, ligand designs require the generated samples with high affinity and specificity that bind to target proteins. Here we presents the most widely used metrics for ligand generation evaluation.
\begin{itemize}
  \item  Drug like linearity (QED): a quantitative estimation of drug-likeness combined with several ideal molecular properties.
  \item[$\bullet$] Synthetic accessibility (SA): a measurement of the accessibility of molecule synthetic.
  \item[$\bullet$] Diversity: the average pairwise Tanimoto difference among molecules generated for each pocket.
  \item[$\bullet$] 3D similarity: the 3D similarity between generated molecules and reference ones, which is calculated by the overlapping ratio of two molecules in 3D space.
  \item[$\bullet$] Vina score: an estimation of bonding affinity between generated ligands and target pockets. This is one of the most important indicators in this research areas.
  \item[$\bullet$] High affinity: the proportion of molecules whose Vina Score is higher than the benchmark molecule in the test set.
  \item[$\bullet$] Lipinski: the ratio of generate ligands that comply with the Lipinski five rule \cite{lipinski_97_lipinski}, which is a rough rule for molecular drug similarity evaluation.
  \item[$\bullet$] LogP: a measurement of the octanol-water partition coefficient. A molecule is considered good drug candidate if its LogP is between -0.4 and 5.6.
  \item[$\bullet$] Mean percentage binding gap (MPBG): it represents the binding affinity of generated molecules on averge.
\end{itemize}

(4) Ligand docking

The ligand docking task involves predicting the position and conformation of a given molecule that binds to the target protein. Models in is particular area are mainly evaluated by RMSD score between generated and reference molecules.

(5) Protein generation

Protein generation includes antigen-specific antibody design, protein sequence and structure deign, and motif-scaffold design. Here we present some of the most commonly adopted metrics. \textbf{Root mean square deviation {RMSD}} in protein generation calculates the distance of $C_\alpha$ between the generated structure and the original structure after alignment. \textbf{Amino acid recovery rate (AAR)} values the fidelity of generated sequences to reference sequences. Another frequently used metric is \textbf{perplexity (PPL)}, which measures the inverse likelihood of native sequences in the predicted sequence distribution.

\subsubsection{Metrics for computer vision tasks} 

(1) Human motion synthesis

The purpose of human motion generation is to synthesize human motion sequences through given information. There are many indicators used to evaluate the synthesis of human motion, and we will elaborate on some commonly used ones.

\begin{itemize}
  \item[$\bullet$] Fréchet inception distance (FID): differences of feature distribution between generated and groud-truth motions.
  \item[$\bullet$] Fréchet video distance (FVD): applications of FID on video sequences.
  \item[$\bullet$] Fréchet gesture distance (FGD): applications of FID on gesture generation.  
  \item[$\bullet$] Diversity: the average joint differences of sample pairs from generated sequences.
  \item[$\bullet$] R-Precision: the measurement of alignment between generated motion and prompts.
  \item[$\bullet$] Multimodality (MM): the measurement of difference among a number of samples generated according to a single input text or action.
  \item[$\bullet$] Multimodal distance (MM Dist): differences between generated motion features and input text features.
  \item[$\bullet$] Kernel inception distance (KID): the skewness, mean, and variance compared in FID.
  \item[$\bullet$] Motion CLIP score (mCLIP): the cosine simlarity between motion and text embeddings from separately trained motion CLIP model.
  \item[$\bullet$] Mutual information divergence (MID): the cross information between motion and text embeddings.
  \item[$\bullet$] Beat alignment scores: the tendency of dance sequences to follow music beat.
  \item[$\bullet$] Average position error (APE): the average positional difference between generated and real motions.
  \item[$\bullet$] Averagge variance error (AVE): the variance of difference between generated and real motions.
\end{itemize}

(2) Human motion prediction

Unlike Human motion synthesis, human motion prediction requires accurate predictions of motions based on observed sequecnes. Evaluation metrics for human motion prediction are more inclined towards error calculation.

\begin{itemize}
\item[$\bullet$] Displacement error (DE): the displacement error of all joints in each frame. There are variation such as average displacement error (ADE), final displacement error (FDE), minimum displacement error (MDE), standard deviation of displacement error (SDE), average final displacement error (AFDE), minimum final displacement error (MFDE), standard deviation of final displacement error (SFDE), Multi-Modal-ADE (MMADE), and Multi-Modal-FDE (MMFDE). 
\item[$\bullet$] Average pairwise distance (APD): L2 distance between among all action examples.
\end{itemize}

(3) Scene graph generation

Performance of scene graph generation methods are evaluated from three aspacts, including scene-level fidelity, scene graph consistency, and object-level fidelity.
FID and KID are used to evaluate scene-level fidelity. 
To evaluate the scene graph consistency, the accuracy of a set of relations between generated layout and scene graph is adopted.
In terms of object level fidelity, Minimum Matching Distance (MMD), Coverage (COV), and 1-Nearest Neighbor Accuracy (1-NNA) are adopted to compute 3D point clouds between generated samples and groud-truth shapes.

\subsubsection{Generic graph generation metrics}
In this subsection, we elaborate on metrics for generic graph generation methods.

(1) Structural evaluation metrics

These metrics are used to assess the structural properties of generated graphs and their alignment with original graphs.

\begin{itemize}
    \item Degree Distribution: Measures the distribution of node degrees in the generated graph, reflecting local connectivity patterns.
    \item Clustering Coefficient Distribution: Measures the distribution of clustering coefficients among nodes in the generated graph, indicating the density of connections among neighbors and clustering tendencies.
    \item Orbit Count Distribution: Measures the occurrence frequency of different orbits in the generated graph, typically focusing on orbits with four nodes.
    \item Laplacian Spectrum: Describes the distribution of eigenvalues of the Laplacian matrix of the generated graph.
    \item Characteristic Path Length (CPL): Measures the average shortest path length between nodes in the generated graph.
    \item Assortativity Coefficient (AC): Measures whether nodes with similar degrees in the generated graph tend to be connected to each other.
    \item Triangle Count: Counts the number of triangles (closed loops of three nodes) in the graph.
    \item Square Count: Counts the number of squares (closed loops of four nodes) in the graph.
\end{itemize}

(2) Quality evaluation metrics

These metrics evaluate the quality of generated graphs and their similarity to the distribution of original data.

\begin{itemize}
    \item Frechet ChemNet Distance (FCD): Evaluates the distance between generated and training graphs in chemical space using activations from the penultimate layer of ChemNet. It assesses the diversity and quality of generated molecules, capturing the similarity in chemical properties between real and generated molecules.
\end{itemize}
\begin{equation}
\text{FCD}(x, y) = \|\mu_x - \mu_y\|_2^2 + \text{Tr}(\Sigma_x + \Sigma_y - 2(\Sigma_x \Sigma_y)^{1/2})
\end{equation}

where \(\mu_x\) and \(\mu_y\) are the mean vectors of the generated and training molecules, respectively, and \(\Sigma_x\) and \(\Sigma_y\) are their covariance matrices. \(\text{Tr}\) denotes the trace of a matrix.

\begin{itemize}
    \item Neighborhood Subgraph Pairwise Distance Kernel (NSPDK) MMD: Computes the Maximum Mean Discrepancy (MMD) between generated and test graphs based on node and edge features (degree distribution, clustering coefficient, orbit count distribution, Laplacian spectrum).
\end{itemize}

\begin{align}
\text{MMD}^2(\mathcal{X}, \mathcal{Y}) = \frac{1}{m^2} \sum_{i=1}^m \sum_{j=1}^m k(x_i, x_j) + \nonumber \\ 
\frac{1}{n^2} \sum_{i=1}^n \sum_{j=1}^n k(y_i, y_j) - \frac{2}{mn} \sum_{i=1}^m \sum_{j=1}^n k(x_i, y_j)
\end{align}

where \(\mathcal{X} = \{x_1, \ldots, x_m\}\) is the sample set of generated graphs, \(\mathcal{Y} = \{y_1, \ldots, y_n\}\) is the sample set of test graphs, and \(k\) is a kernel function, commonly a Gaussian kernel.

\begin{itemize}
    \item Generation Time: Measures the time required to generate a specified number of graphs.
\end{itemize}

\subsection{Datasets}
\subsubsection{Datasets for researh on AI for scientific discovery}
ZINC \cite{zinc_12_irwin} is an excellent virtual screening compound library that contains over 750 million purchasable compounds, 230 million compounds of which are in 3D format. GEOM-QM9 \cite{qm9_14_ramakrishnan} and GEOM-Drugs \cite{drugs_22_axelrod} are the most commonly used datasets for 3D molecule and conformation generation. The former one contains molecule conformations in equilibrium state, and is a widely used benchmark dataset for property predictions of molecules in equilibrium state. It contains approximately 133000 molecules, with up to 9 heavy atoms or 29 atoms with hydrogens included.There are only 5 types of atoms: H, C, O, N, and F. GEOM-Drugs \cite{drugs_22_axelrod} includes 37 million molecular conformations for more than 450000 molecules. The maximum number of atoms in this dataset is 181 while the averge is 44. CrossDocked2020 \cite{crossdock_20_paul} is a dataset of protein-ligand complexes, and known for the inclusion of complexes with both high and low affinity. It contains over 22 million protein-ligand crystal structures. BioLiP \cite{biolip_12_yang} is a database of protein-ligand interaction collected from multiple sources get updated weekly. Till June 2023, BioLiP includes over 800000 complexes, over 450000 ligands, and over 400000 proteins. PDBbind \cite{pdbbind_17_liu} is composed of protein-ligand complexes from Protein Data Bank (PDB). The latest version of it contains over 19000 protein ligand structures. SAbDab is a database composed of all the antibody structures available in the PDB, the total number of antibody structures in SAbDab in the latest version is 7324. For protein generation tasks, PDB provides access to 3D structures of proteins, nucleic acids, and ligand-protein complexes. 
Perov-5 \cite{perov51_12_castelli,perov52_12_castelli} contains perovskite structures with different compositions.
Carbon-24 is a dataset of carbon materials simulated 10GPa and MP-20 \cite{mp_13_jain} is a collection of experimentally-generated materials from Material Projects.
The most frequently used datasets for molecule generation are summarized in Table~\ref{tab:data_molgen}.

\begin{table*}
  \centering
  \caption{Datasets for Diffusion-based Graph Generation}
  \label{tab:data_graphgen}
  \begin{tabular}{ccc}
    \hline
    \textbf{Dataset} & \textbf{Description} & \textbf{Task} \\
    \hline
    ZINC & 250,000 molecules & De nove molecule design, generic graph generation\\
    GEOM-QM9 & 133K molecules and conformations & \makecell[c]{De novo molecule design, conformation design, \\ generic graph generation} \\ 
    GEOM-Drugs & 450K molecules and 37M conformations & De novo molecule design, conformation design \\ 
    CrossDocked2020 & \makecell[c]{13.8K ligands, 2.9K pocktes, \\18K complexes, and 22.6M poses} & De novo ligand design \\ 
    BioLiP & 450K ligands, 400K pockets, 800K complexes  & De novo ligand design \\ 
    PDBBind & 19K complexes & De novo ligand design, ligand docking \\ 
    SAbDab & 7K antibody structures & Protein generation \\
    Perov-5 & 19K perovskite materials & Crystal structure prediction \\
    Carbon-24 & 10K carbon materials & Crystal structure prediction \\
    MP-20 & 45K inorganic materials & Crystal structure prediction \\
    Community & 500 community graphs & Generic graph generation \\
    Enzymes & 578 protein graphs & Generic graph generation\\
    Grid & 100 2D grid graphs & Generic graph generation \\   
    Human3.6M & 3.6M poses, 11 subjects & Human motion synthesis, human motion prediction \\ 
    HumanEva-I & \makecell[c]{7 sequences, 20.6K frames, \\6 action classes from 4 subjects} & Human motion synthesis, human motion prediction \\ 
    HumanAct12 & 1.2K sequences, 90K frames, 12 subjects & Human motion synthesis \\ 
    HumanML3D & 14.6K sequences, 45K descriptions & Human motion synthesis \\ 
    KIT &  3.9K sequences, 6.5K descriptions & Human motion synthesis \\ 
    BABEL & 13K sequences, 63k labels, 252 subjects & Human motion synthesis \\ 
    UESTC & 24K sequences, 40 subjects & Human motion synthesis \\ 
    3DPW & 60 sequences, 51K, 5 subjects & Human motion synthesis \\
    NTU RGB+D 60 & \makecell[c]{57K sequences, 60 action classes\\from 40 subjects} & Human motion synthesis \\
    NTU RGB+D 120 & \makecell[c]{114K sequences, 120 actions classes\\from 106 subjects} & Human motion synthesis \\ 
    AIST++ & 1.4K sequences & Human motion synthesis \\ 
    TSG & 200 sequences, 16.2M frames, 50 subjects  & Human motion synthesis \\
    ZeroEGGS & 67 sequences, 8K frame, 19 subjects & Human motion synthesis \\
    3D-FRONT & 6.8K CAD houses, 18.7K furnished rooms, 7.3K furniture objects & Scene graph generation \\
    SG-FRONT & 5,7K indoor scenes from 3 scene types with 45K objects & Scene graph generation \\
    \hline
\end{tabular}
\end{table*}


\subsubsection{Datasets for research on computer vison}
Human3.6M \cite{human36m_14_ionescu}, which is composed of 3.6 million human poses divided into 15 categories, is the largest benchmark dataset for human motion studies. Originally, human skeletons in it has 32 joints, but they are not completely adopted in experiments. AMASS \cite{amass_19_mahmood} is a combination of 18 datasets, which makes it a large one contains 13944 motion sequences. HumanAct12 \cite{humanact12_20_guo} is a dataset derived from PHSPD \cite{sfp_20_zou,phsps_20_zou}. It has 1191 sequences and 90099 frames splitted into 12 categories in total. By re-annotating motions in AMASS and HumanAct12, HumanML3D \cite{humanml3d_22_guo} is a motion-text pair dataset consisting of 14616 sequences with 44970 descriptions. KIT \cite{kit_16_plappert} is also a dataset with natural language annotations. Based on AMASS, BABEL \cite{babel_21_punnakkal} contains 13220 sequences seperated into 66289 segments. The dataset is organized into 252 subjects and annotated with 63353 frame-level labels. Being the first 3D motion dataset in the wild, 3DPW \cite{3dpw_18_marcard} has 60 sequences, and 27 two-person sequences annotated with SMPL joints. UESTC \cite{uestc_18_ji} contains 24K sequences with 40 subjects. NTU RGB+D \cite{nturgbd_16_shahroudy} is a widely used dataset that is often adopted in the form of NTU RGB+D 60 and NTU RGB+D 120, depending on the number of subjects included. AIST++ \cite{aist++_21_li} is a dataset with high-quality dance motions with 10 different dance motion genres. TSG \cite{tsg_18_ferstl} contains 4 hours of full body motion with speech audio, while ZeroEGGS \cite{zeroeggs_23_ghorbani} is consist of two hours of motion and audio data in 19 styles. These two datasets are usually adopted in gestrue genration tasks. 3D-FRONT \cite{3dfront_21_fu} is a collection of synthetic indoor scene furnished with high-quality objects without detailed discription. Based on 3D-FRONT, SG-FRONT \cite{commonscenes_23_zhai} provides annotation labels on a number of object relationships.
Details of the datasets are listed in Table~\ref{tab:data_graphgen}.


\section{Future directions}
Though diffusion-based graph generative methods achieve huge progress recently, there still remains some challenges to be tackled. 

\textbf{Training objectives and evaluation metrics:} Training objectives of current implementations of diffusion methods on graphs still takes the form of evidence of lower bound (ELBO) of negative log-likelihood. It has not been proven that optimizing ELBO and the negative log-likelihood are equivalent. Several studies \cite{midi_23_vignac,mdm_22_tevet} design loss terms specifically for different tasks, but they are not widely adopted in other research. There is plenty of room for further explorations. Meanwhile, as research attention on graph generation shifts from 2D to 3D geometry graphs, many metrics used for 2D graph generation are not sufficient enough to comprehensively evaluate the quality of generated samples. New evaluation metrics that reflect the quality of generated geometry structures are in need. 

\textbf{The disrete nature of Graphs:} Though previous diffusion-based methods mainly focus on continuous data, there are many research that tackles the discrete nature of graphs by many approaches. Most research deal with this problem by directly diffusing one-hot feature vectors or adjacency matrices \cite{edpgnn_20_niu}. While there are several studies \cite{digress_22_vignac,midi_23_vignac} make attempts to adapt diffusion on discrete graph data by introducing new probabilistic distributions or mapping discrete data into continuous space. To summarize, there is no universal approach for graph diffusion and edge generation.

\textbf{Application fields:} Most applications of generative graph diffusion methods are on molecule and motion generation. This leaves applications on other fields rarely explored. Though there are few research that focus on graph structure predictions and spatio-temporal graph generation, applications on popular graph learning tasks, such as recommender systems and anomaly detections, receive minor research attention. Studies \cite{dpmgsp_23_jang,diffstg_23_wen} show that diffusion models have great potential in supervised learning and spatio-temporal graph forcasting. This shed a light upon potential applications on sequential recommendations, traffic prediction, and weather forecasting.

\textbf{Out-of-distribution and context-aware generation:} In practical research, models often suffer from diverse data sources and distribution shifts. Such out-of-distribution problems may lead to degradation of performance when models are adapted to emergent datasets. However, only a few efforts are devoted to this problem. For example, MOOD \cite{mood_22_lee} tackles the problem by introducing a OOD-controlled diffusion process. Further exploration in the sampling phase, context-aware capacity, and denoising kernel could bring new insights to OOD problems. Generating samples with desirable properties is the ultimate goal of many genration tasks. For motion generation with given contextual information, this goal is easier to achieve. However, when it comes to molecule generation that requires more physical and chemical knowledge, further studies are in need to endow models with the ability to generate samples with expected characteristics.

\textbf{Combination with AIGC:} Recent advances in artificial intelligence generated content (AIGC) \cite{stablediffusion_22_rombach,dalle_21_ramesh} arouse widespread research attention and enthusiasm. The pre-training and fine tuning structure of GPT \cite{gpt3_22_long,gpt4_23_openai} displays promising abilities to generate high-quality samples that are tough for human to distinguish. Combining diffusion-based models with AIGC may result in even better performances.

\section{Conclusions and discussion}
\subsection{Conclusions}
Being the most advanced generative methods, diffusion-based methods achieved great progress in graph generation tasks, especially on molecule design and motion synthesis. In this paper, we make a comprehensive review on diffusion-based graph generative methods. We first analyze different paradigms of diffusion methods and their applicaitons on graphs. Then, applications on various tasks and prevailing datasets and metrics are elaborated. As far as we are aware, this is the most comprehensive survey on diffusion-based graph generative methods so far.

\subsection{Comparison with Other Surveys}
In the field of diffusion based graph generative methods, surveys by Zhang et al. \cite{graphdiffsurvey_23_zhang} and Liu et al. \cite{gendiffgraph_23_liu} share some similarities with our survey. Both their survey mainly focuses on applications of diffusion-based graph generative methods on AI for science discovery. We further elaborate on the applications on computer vision and provide a brief summary on datasets and metrics. We also provide detailed discussion on research that addresses the anisotropic nature of graphs and enhances adaptability of diffusion on graphs.

\bibliographystyle{IEEEtran}
\bibliography{mybibfile}


 





\end{document}